# Toward Task Capable Active Matter: Learning to Avoid Clogging in Confined Collectives *via* Collisions

***Kehinde O. Aina*[1], *Ram Avinery*[2], *Hui-Shun Kuan*[3], *Meredith D. Betterton*[3,4], *Michael A. D. Goodisman*[5] and *Daniel I. Goldman*[2]***

*[1]Institute for Robotics and Intelligent Machines, Georgia Institute of Technology, Atlanta, GA, United States, [2]School of Physics, Georgia Institute of Technology, Atlanta, GA, United States, [3]Department of Physics, Department of MCD Biology, University of Colorado Boulder, Boulder, CO, United States, [4]Center for Computational Biology, Flatiron Institute, New York, NY, United States, [5]School of Biological Sciences, Georgia Institute of Technology, Atlanta, GA, United States*

***Correspondence:**
*Daniel I. Goldman*
*daniel.goldman@physics.gatech.edu*

Social organisms which construct nests consisting of tunnels and chambers necessarily navigate confined and crowded conditions. Unlike low density collectives like bird flocks and insect swarms in which hydrodynamic and statistical phenomena dominate, the physics of glasses and supercooled fluids is important to understand clogging behaviors in high density collectives. Our previous work revealed that fire ants flowing in confined tunnels utilize diverse behaviors like unequal workload distributions, spontaneous direction reversals and limited interaction times to mitigate clogging and jamming and thus maintain functional flow; implementation of similar rules in a small robophysical swarm led to high performance through spontaneous dissolution of clogs and clusters. However, how the insects learn such behaviors and how we can develop "task capable" active matter in such regimes remains a challenge in part because interaction dynamics are dominated by local, potentially time-consuming collisions and no single agent can survey and guide the entire collective. Here, hypothesizing that effective flow and clog mitigation could be generated purely by collisional learning dynamics, we challenged small groups of robots to transport pellets through a narrow tunnel, and allowed them to modify their excavation probabilities over time. Robots began excavation with equal probabilities to excavate and without probability modification, clogs and clusters were common. Allowing the robots to perform a "reversal" and exit the tunnel when they encountered another robot which prevented forward progress improved performance. When robots were allowed to change their reversal probabilities via both a collision and a self-measured (and noisy) estimate of tunnel length, unequal workload distributions comparable to our previous work emerged and excavation performance improved. Our robophysical study of an excavating swarm shows that despite the seeming complexity and difficulty of the task, simple learning rules can mitigate or leverage unavoidable features in task capable dense active matter, leading to hypotheses for dense biological and robotic swarms.

**Keywords: collision-based interaction, collective behavior, multi-robot excavation, swarm robotics, decentralized learning, ant-inspired learning, active matter, confined and crowded conditions**

# 1 INTRODUCTION

Active matter systems, ensembles of driven "agents", are of much interest in physics for their rich phenomena, which often feature formation of spatially extended structures such as those observed in flocking [1, 2], motility-induced-phase-separation [3–5], giant number fluctuations [6–8] and more [6, 9]. Active systems in confined environments (like within narrow channels) are interesting as structures that form due to collisions and constrained maneuverability in this regime [10–13] decay slowly in time, displaying glassy/supercooled features [11]. Such slow relaxation can lead to deleterious performance of an active system which must perform a "task" (like a group of ants, termites creating tunnels or humans rushing through narrow doors). General physics principles which could allow such active systems to become "task capable" are less understood as the bulk active matter physics focuses on flow and structures that emerge from relatively simple rules among individuals.

Mitigating structure formation in confined systems likely necessitates agents change behavior ("rules") in response to conditions and via interactions with other agents. Studies of such systems are typically the domain of swarm engineering where researchers seek to understand the functional benefits of structure formation. For example, engineers seek to have robot teams achieve goals such as getting aerial swarms to create formations [14] or planar collections of robots to arrange in different patterns [15]. Generically, swarm control schemes may modify a steady-state property of the system such as cluster size [16], pattern formation [17] and locomotion alignment [18]. Such schemes are often represented as functional dependencies between variables, like the orientation being the average of neighbor orientations [1] or the speed decreasing with local particle density [5].

Unfortunately, most control schemes for swarms assume dilute conditions and avoid collisions thus discovery of general principles for task completion (like flow at high speed or low energetic cost) in crowded confined conditions requires new insights in part because real-time adjustment is particularly challenging without a central controller, and with the limited sensing and computation we imposed on the robots, all while dealing with physical noise from their mechanics, jostling and collisions. As a result, conventional planning and control methods that rely on precise or accurate information of the surrounding may not be applicable to achieve coordinated behavior and good traffic flow in such a setting. It is instead useful to discover decentralized learning rules that rely on the unavoidable features of these dense active systems–social and local interactions–to reach effective traffic flow and task performance, under evolving conditions. We wish to understand then broadly how such structures form or dissolve in collectives whose agents possess memory, sensory feedback and even capability to learn over time.

Ants and termites are biological examples of dense and crowded task oriented active systems, where various behaviors (i.e., control schemes) have been naturally selected to aid task performance in such regimes [19, 20] without central control. Ants, for example, cooperatively create nests with complex subterranean networks [21]. They employ no centralized controller or global information, yet are able to excavate soil in dark, narrow and overpopulated conditions [22, 23]. Their tasks usually involve manipulation of soil particles or substrates, transport of bulk pellets through long and narrow tunnels, as well as directed movement to and from their nests [24]. Controlled lab experiments and numerical simulations show that clustering and clogging are prevalent in these conditions [11, 12], similar to "glassy arrests" in non-living active matter [25, 26]. This is due to individuals' persistence in their goals, being unaware or inconsiderate of others', which may lead to clogs that are difficult to resolve when working in narrow, quasi one-dimensional tunnels [12].

Previously we used robots as a robophysical model of the ant tunneling system which facilitated testing of behavioral rules in a controlled environment with noise and complexities of the real physical world [12]. Using the robophysical model we demonstrated how an active confined crowded robot collective could mitigate structure formation (slowly dissolving clogs and jams) via being "lazy" and "giving up". That is, manipulations of workload distribution in this robophysical model collective rationalized our observations of biological ants' strategy unequal workload distributions and probabilistic yielding to oncoming traffic (termed "reversals") demonstrating the importance and utilty of such rules for maintaining optimal tunnel flow. However, we had to program the behaviors in the robots; here we are interested to learn how robots can adjust their behaviors to optimize their workload distributions and retreat behaviors thus providing insight into biological collectives as well as providing principles for robot swarms that must operate in crowded, confined conditions.

Therefore to discover principles by which confined swarms can learn to avoid clogging while performing a useful task (excavation), purely via local information, social interactions and a noisy estimate of their state, here we augment our robophysical swarm to investigate hypotheses for how ant encounters can regulate [27–29] activity by individually learning from collisions. We systematically study the performance of the robots as we subject them to different protocols. We show that clogging can indeed be mitigated, by some individuals learning to "give up" and participate less in digging, where social interactions such as inter-robot collisions and noisy estimates of tunnel length serve as a means of reinforcement in our learning scheme. Our learning scheme provides a robust response to changing conditions, even when individuals acquire noisy or inaccurate information about their environment (tunnel length). We expect that our robophysical model and learning technique will provide guidance for biological hypotheses, as well as inform the design of a robust coordination technique for dense swarms in dynamic and evolving environments. Our results provide an example of the richness of active matter dynamics when the agents can use information to change state to perform and learn to perform tasks.

# 2 METHODS

Our robots, modified from our previous study [12] are programmed to execute autonomous behaviors independently such as navigation to specific sections in the tunnel and excavation. They are equipped with force sensitive grippers for pellet excavation, an outer shell with capacitive sensing to detect and distinguish two types collisions - robot-robot collisions and robot-wall collisions, as well as terminal rods for charging and detecting the home area (**Figure 1**). The pellets to be excavated are laid at the end of the tunnel are a cohesive granular medium consisting of plastic shells housing loose rare-earth magnets.

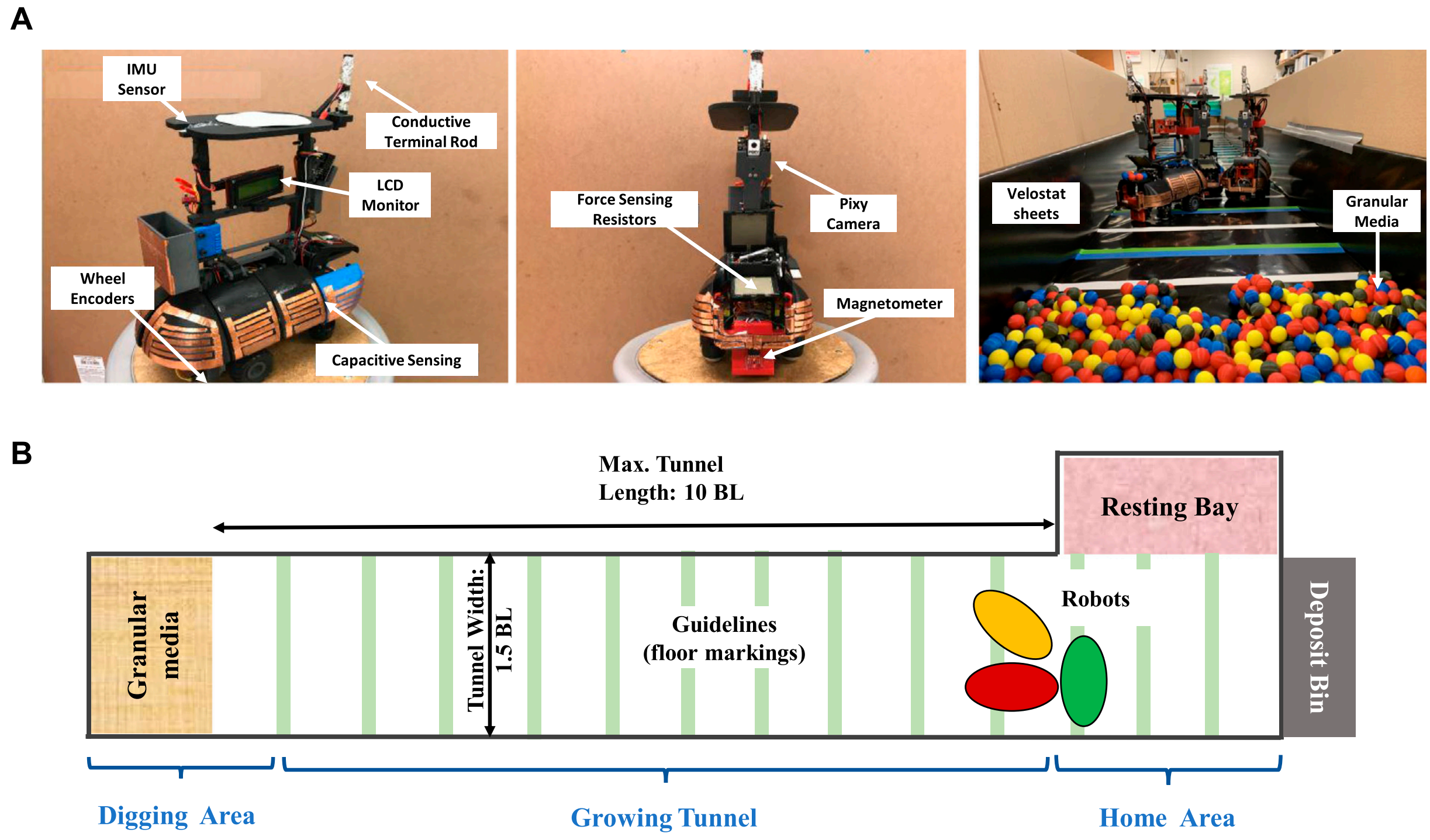

**FIGURE 1 |** Experimental apparatus consisting of robots and a tunnel. **(A)** Components of an ant robot (two left images): Inertia Measurement Unit (IMU) sensor for absolute orientation; Wheel encoders for localization via robot odometry/dead reckoning; Capacitive touch sensor shells for detecting collisions; Force sensitive resistors for gripper tactile feedback, Magnetometer for detecting pellets and Digging Area; Pixy camera for tunnel navigation and localization; Terminal rod for charging and detection of Home area. In the tunnel (rightmost image): Velostat sheets for lining the tunnel to distinguish robot-robot collision (copper-copper) from robot-wall collision (copper-velostat); Granular media to simulate model cohesive soil excavated by ants **(B)** Schematic diagram of the confined tunnel: a robot starts from the Home Area, decides whether to enter the tunnel and excavate some pellets, or decides to go to the Resting Bay to avoid interference at the tunnel entrance.

Since we are particularly interested in how coordinated group success could be achieved only from physical interactions and local observation of the environment, we do not allow direct robot-to-robot communication or global information to the individuals in the group; our robots rely purely on on-board sensors and make decision based on local sensing and self-reinforcement.

## 2.1 Collective Task

Our task is a collective excavation scenario: a group of robots must continuously excavate the model granular media in a narrow (1.5 body lengths) and confined tunnel, shown in **Figure 1**. As more pellets are excavated, the tunnel "grows" or changes geometrically as the robots perform their task. A robot starts by leaving the Home area and, using vision, following the guiding trails to the digging area where the cohesive pellets are located. During transit in the tunnel, the robot can detect and distinguish collisions with other robots, as well as collisions with the wall of the tunnel. By sensing a magnetic field, the robots can also detect the pellets. After a successful attempt to excavate, a robot heads home to drop the excavated pellets into a Deposit bin placed on a weighing scale.

Our goal is for the group to excavate as many pellets as possible within a given time. An obvious solution is for the robots to remain constantly active and try to excavate; however as we demonstrated in [12] when all the robots are in such a mode and enter the tunnel concurrently they spend much of their time resolving collisions as a result of competition for space to maneuver and carryout their activities. The resulting traffic jams have robots stuck or stalled due to excessive stress from repeated collisions. This wasted time results in degraded performance, to the extent that fewer robots in the tunnel would excavate faster. The challenge is therefore to use the local information available to individual robots to regulate the congestion and improve group performance under such physical constraints and hindrances. We derive our inspiration from the social behavior of fire ants under crowded and confined conditions [12], and our previous robophysical-model excavation experiment [30], to develop an adaptive learning rule that makes the robots decide when to "give up" digging and when to "take a rest" in a way that significantly improve the performance of the group.

## 2.2 Robot Controller

We adopt a finite state automaton model [31] which is a common scheme used to control behavior-based robot activities with no global knowledge. A state transition is triggered when a robot senses some physical clues from the environment. Each sensor on the robots has a specific trigger state that enables the robot to transition into another state. **Figure 2** shows the model of individual robot's controller. Each block contains a set of states or sub-states that form a mode or behavior that the robot exhibits. The states and sub-states are as described below:

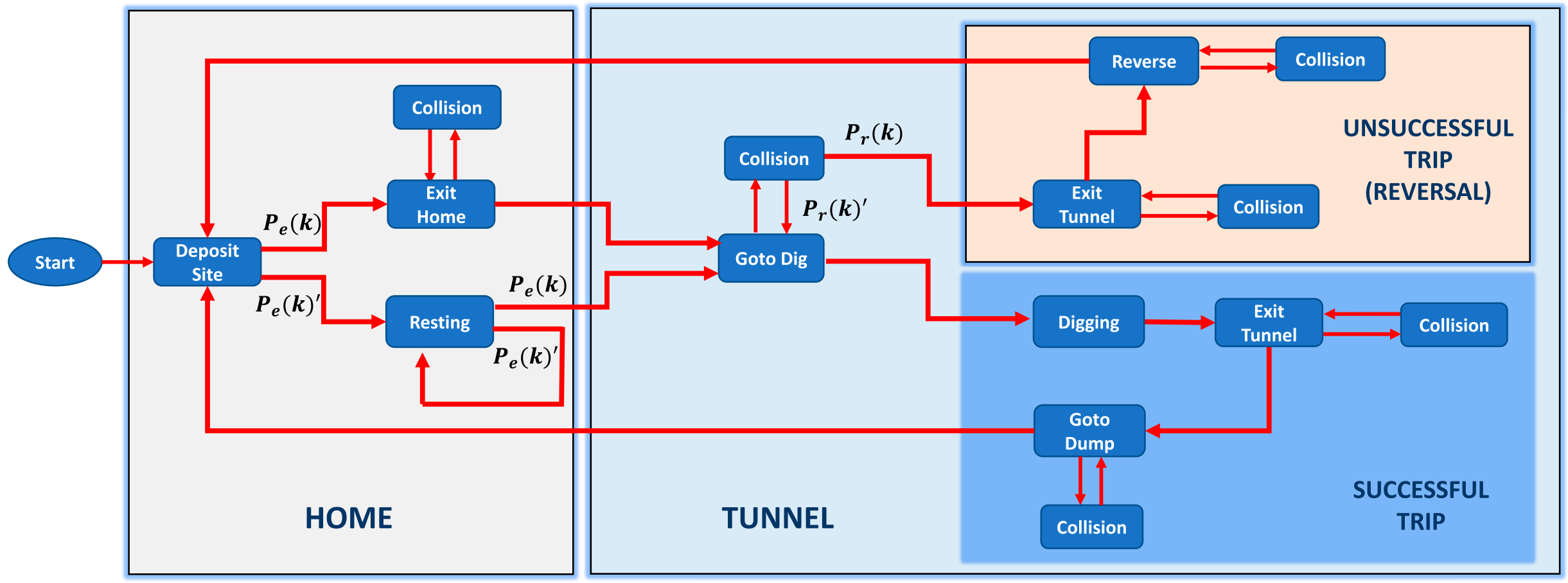

**FIGURE 2 |** Block diagram for the individual robot controller. State transitions are event-based. $P_e$: probability of tunnel entry (or go to dig). $P_e'$: probability of resting (or staying at home). $P_r$: probability of "giving up" after colliding with another robot in the tunnel. $P_r'$: probability of continuing to dig after colliding with anther robot.

1. Goto Dig: This state is triggered at the start of each trip when a robot "decides" whether to dig based on their tunnel entrance probabilities $P_e(k)$.
2. Digging: This state is triggered when the robot is in proximity to the granular media. The magnetometer at the base of the robot detects the magnetic field of the granular media and prompts the robot to start the excavation routine.
3. Exit Tunnel: The robot enters this state from Digging when the Force Sensitive Resistor (FSR) detects sufficient amount of pellets in the gripper. The robot executes several turning maneuvers to exit the digging area and head back home.
4. Goto Dump: This is the state that captures the robot heading home after a successful pellet retrieval (Successful Trip) or unsuccessful pellet retrieval (Unsuccessful Trip). The controller drives the robot out of the tunnel and gets the robot home to the deposit area.
5. Dumping: Robot releases the excavated pellets from its gripper and dumps it in the "Deposit Bin" which is placed on a weighing scale to measure the amount of pellets excavated over time.
6. Exit Home: Robot executes some turning maneuvers to exit the deposit area and enters the tunnel to dig.
7. Collision: This state is triggered when a robot collides with another robot or with the tunnel wall. The robot executes a set of turning maneuvers in an attempt to resolve the collision.
8. Resting: Robot goes to this state at the beginning of a trip if the entrance probability is sampled and the robot decides to rest. The robot follows the guiding trail on the tunnel floor to navigate to the resting area to take a rest and not participate in the tunnel traffic.

We developed a stochastic model with two parameters to control the entrance rate and reversal rate (give-up rate) of the robots so as to regulate tunnel traffic and improve group performance. Let $P_e(k)$ be the tunnel entrance probability and $P_r(k)$ be the reversal probability of each robot at trip attempt number $k$. A trip begins when a robot samples from the entrance probability, $P_e(k)$ and decides whether to "go in and dig" or "stay at home and rest". This parameter controls the number of robots in the tunnel which directly controls the tunnel density or congestion rate. The reversal probability, $P_r(k)$, on the other hand controls how a robot responds to a collision when it occurs. A robot samples from this parameter and decides if it should "give up" or to continue its journey. With these two parameters, we developed two protocols for studying the effects and performance of fixed social behaviors [12] in multi-robot collective excavation. We use these previously reported fixed behavior protocols [12] as controls to test against the adaptive (learning) behaviors we develop in the next section:

Active Protocol: In this protocol, we fix the tunnel entrance probability $P_e(k)$ to a value of one for each trip for each robot. The reversal probability $P_r(k)$ is set to zero, so the robots do not return home until they are able to collect pellets. This ensures that all the robots are active, trying to dig in the tunnel at all times.

Reversal Protocol: Here we set the reversal probability $P_r(k)$ for each robot to a value greater than zero but less than one, while still keeping the entrance probability to one at all time. This allows the robots to randomly "give up" trying to dig when they collide with other robots in the tunnel.

## 2.3 Developing an Adaptive Protocol

To go beyond the above fixed behavior protocols and to gain insight into useful social interactions of the confined multi-robot system, we conducted a parameter sweep to find the optimal reversal probabilities that yielded the highest excavation rate in a Cellular Automata (CA) model developed in our previous work [12]; see **Supplementary Section** for detailed description of the model. **Figure 3A** shows the range of excavation rates for varying reversal probability as the tunnel length increases. A closer look at the region with highest excavation rates suggests a non-linear inverse relationship between optimal reversal probabilities and tunnel length. In particular, the optimal probability values drop sharply for short tunnels (less than 5BL) and more gradually for longer tunnels. This gives us inspiration to develop an Adaptive

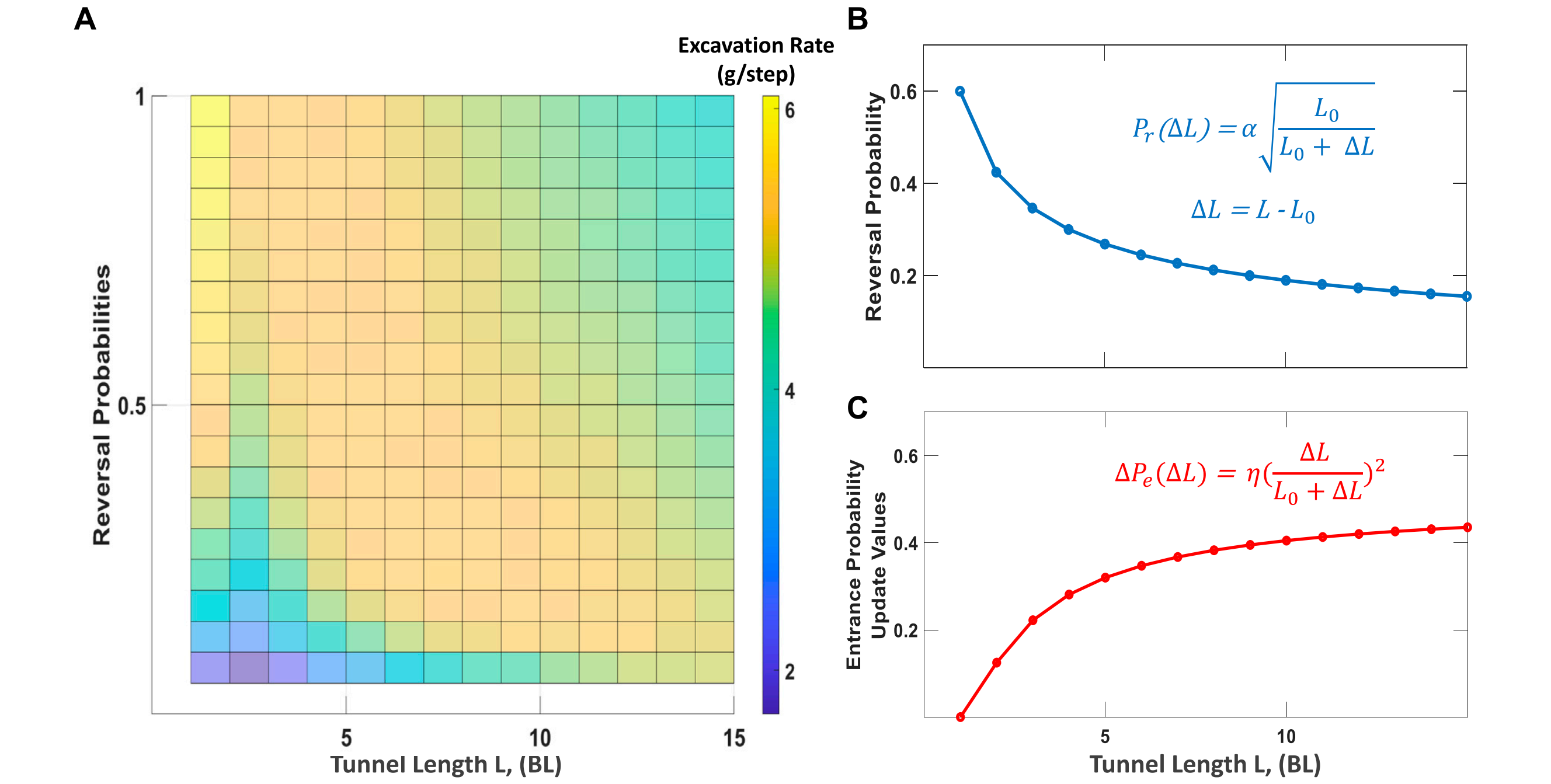

**FIGURE 3 |** Probability sweep over tunnel length with selected equation models. **(A)** Reversal probability sweep vs. Tunnel length. Plot shows that high to medium reversal probabilities give the highest excavation rate for short tunnels, while low reversal probabilities give the highest excavation rates for long tunnels **(B)** Selected equation for modelling Reversal probabilities as a function of tunnel length for multi-robot experiments **(C)** Learning update value for Entrance probabilities as a function of tunnel length for multi-robot experiments.

Reversal probability function, $P_r(\Delta L)$, that decreases sub-linearly as the tunnel length grows (**Figure 3B** and **Eq. 1** below).

Additionally, ant excavation studies and CA model analysis reveal that an asymmetric or unequal workload distribution improves excavation performance in confined conditions [12, 24]. To attempt to incorporate this principle/strategy in our robophysical system, we develop a self-reinforcement protocol where we model the digging desire of individual robots with a probability $P_e$, called the "tunnel entrance probability", and update its value based on if a digging attempt performed by a robot was successful (increase $P_e$ by a constant), or unsuccessful (decrease $P_e$ by a constant). Preliminary results showed that this protocol consistently produced an unequal workload distribution with better excavation performance than the Reversal protocol in long tunnels (~10BL). However, this was not the case for short tunnels. Unequal workload strategy did not perform better when the tunnel was short ( $<$ 4BL). To account for this phenomenon, we crafted an adaptive "update value" function, $\Delta P_e$, (**Figure 3C** and **Eq. 2** below, which has a small update value when the tunnel is short. We call this the Adaptive Protocol. That is, our adaptive protocol modifies the digging desire of individual robots by using egocentric estimates of the change in tunnel length to update individuals' entrance probability values.

Intuitively, the Adaptive protocol (*via* **Eqs 1**, **2**) suggest that the cost of "giving up" due to collisions (high density) at longer tunnels is substantial, and the strategy to minimize congestion is by deploying fewer workers to dig, or equivalently, more workers to rest. That is, long-duration clogging is more likely to occur in longer tunnels than in short tunnels due to the cascaded effects of multi-body collisions propagated as the robots or ants flow into the tunnel simultaneously. Our adaptive protocol addresses this issue by having the robots modify their entrance probabilities slowly first at the initial stage of digging, then more rapidly at the later stage (**Eq. 2**). $\xi$ is a parameter added to ensure that a resting robot does not remain in resting mode indefinitely (i.e. that $P_e$ does not go to zero) allowing robots to explore their environment, update their estimates of the change in tunnel length, and modify their behaviors if necessary.

$$P_r(\Delta L) = \alpha\sqrt{\frac{L_0}{L_0 + \Delta L}} \tag{1}$$

$$P_e(k, \Delta L) = \begin{cases} P_e(k-1) + \eta\Delta P_e(\Delta L) & \text{if successful trip} \\ P_e(k-1) - \eta\Delta P_e(\Delta L) & \text{if unsuccessful trip} \\ P_e(k-1) + \xi & \text{otherwise (resting)} \end{cases} \tag{2}$$

where:

$$\begin{aligned} \Delta P_e(\Delta L) &= \text{entrance probability update value} \\ L_0 &= \text{initial tunnel length (in robot body lengths)} \\ \Delta L &= \text{change in tunnel length} \\ \xi &= \text{noise or exploration term} \\ \alpha &= \text{normalizing constant for reversal probability} \\ \eta &= \text{normalizing constant for tunnel entrance update} \end{aligned}$$

and $\Delta P_e(\Delta L) = (\frac{\Delta L}{L_0+\Delta L})^2$.

The power law expressions of **Eqs 1**, **2** are simple forms that yield the desired behaviors, i.e. rapid change in the reversal and update values at short tunnels and slow/gradual change at long tunnels (**Figures 3B,C**). Other power law exponents can also be used and will likely result in various degrees of performance gains, as long as the exponent is $< 1$ for the reversal probability and $> 1$ for entrance probability update. **Eq. 2** takes a positive sign if the robot is able to get pellets home (successful trip), otherwise it takes a negative sign (unsuccessful trip).

Each time a robot reaches the digging area and excavate pellets, it updates its estimate of the change in tunnel length, $\Delta L$ as follows:

$$L(k) = L(k-1) + \gamma\big(L(k)' - L(k-1)\big) \tag{3}$$

$$\Delta L = L(k) - L_0 \tag{4}$$

where:

$L(k)'$ = new measurement of the tunnel length, derived from robot odometry (or dead – reckoning) using the wheel encoder readings.
$L(k)$ = estimate of the tunnel length, averaged over old and new measurements.
$L(k-1)$ = updated value of the tunnel length during the last successful trip
$\gamma$ = the weighting parameter or learning rate.

A new measurement of the tunnel length, $L(k)'$, is computed when a robot successfully reaches the digging site and excavates pellets. At this time, the tunnel length is derived from the x-component of the robot's location as computed by the robot odometry [32]. The robots use the kinematic model of a differential drive mobile robot based on wheel encoder counts to estimate their absolute displacements in the tunnel. The derivation is provided in the **Supplemental Section**. **Eq. 3** above is an exponential moving average formula that acts as a filter for the estimate of the tunnel length which is used to compute $P_r$ and $P_e$. It has an important application of reducing noise in a robot's estimate of the tunnel length which might occur when the robots are in multiple collisions. We chose our value of $\gamma$ to be 0.9 which results in good performance for our experiments. Each robot maintains a separate copy of the equations and updates $P_e(k)$, $P_r(k)$ and $\Delta L$ asynchronously according to Algorithm 1 described below.

**Algorithm 1:** Adaptive Learning Rule Pseudocode.

```
Initialize: k=1, T_r = 60, P_e(k) = 1.0 , P_r(k) = α;
Set experiment duration, T ;
while t < T do
    Sample p ~ U(0,1) ;
    if p < P_e(k) then
        Goto dig (Active Mode);
        if Contact with a robot then
            Sample q ~ U(0,1) ;
            if q < P_r(k) then
                Exit tunnel (Give up);
            else
                Continue going to dig;
            end
        else
            Continue going to dig;
        end
        if Robot grabs pellets then
            L(k) = L(k-1) + γ(L(k)' - L(k-1)) ;
            ΔL = L(k) - L_0 ;
            Exit tunnel;
        else
            Continue going to dig;
        end
        if Robot gets home with pellets then
            P_e(k) = P_e(k-1) + η(ΔL/(L_0+ΔL))^2;
        else
            P_e(k) = P_e(k-1) - η(ΔL/(L_0+ΔL))^2;
        end
        P_r(k) = α sqrt(L_0/(L_0+ΔL)) ;
    else
        Rest for T_r seconds (Resting Mode);
        P_e(k) = P_e(k-1) + ξ;
    end
    k = k + 1;
end
```

The complexity of Algorithm 1 is proportional to the total number of states, $S$, that the robot visits during an excavation trip. This is denoted as $O(S)$ using the big-O notation. The best-case scenario occurs when the robot does not encounter any collisions but travels from the home area to the digging site and back with pellets. This is likely to occur when there are a few robots in the tunnel and the time to complete a trip (i.e., one pass of the algorithm) is relatively short. On the other hand, the worst-case scenario occurs when the robot encounters and handles collisions in all the states, since collision handling is considered an "intermediate" state (see **Figure 2**). In either case, the amortized run time complexity of the algorithm is $O(1)$ since the number of possible states is bounded and does not depend on any input. Similarly, the space complexity is $O(1)$ since the memory space is fixed and does not grow or depend on any input.

## 3 EXPERIMENT

We implemented the Adaptive protocol on our physical robots to compare its performance with the Active and Reversal protocols. Unlike in our previous robophysical experiments [12] in which tunnel length and digging probabilities did not change during excavation, here we conducted experiments in which the tunnel increased in length as the robots excavated the granular media (pellets). This both better models growth of tunnels in biological collective excavation [12, 21, 24] and demonstrates how our learning scheme can adapt to dynamic and non-stationary environments. **Figure 4** shows three snapshots of the robot experiment setup.

Because of limitations in the robot's excavation performance per trip, we conducted experiments in the following scheme: initially, the granular media was positioned at one body length in the tunnel ($L_0 =$ 1). To model a tunnel increasing in length, the pellets were moved backwards incrementally by one body-length each time the robots made a cumulative deposit increment of 300 g (a camera positioned above a weighing scale recorded the weight of total pellets excavated by the robots). The robots can estimate the tunnel length with a calculation of distance traveled as reported by their wheel encoders (**Figure 1**), and update their reversal and entrance probabilities according to **Eqs 1**, **2** respectively.

At the start of each trial and for each protocol, the entrance probability of the individual robot is set to 1. This ensures that all robots are active and will thus interact with the environment. For the Active and Reversal protocols, the entrance probability remains fixed throughout the duration of the experiment, while for the Adaptive protocol, the entrance probability changes approximately as the inverse square of the tunnel length (**Eq. 2**). This update rule ensures that the robots become less active as the tunnel length increases. Hence, the workload should go from equal to unequal. The $\eta$ parameter is chosen to ensure that unsuccessful robots decide to rest more often when the tunnel is long, so as to not hinder the performance of the robots that can reach the digging area. If a robot samples from the entrance probability and decides to rest, it navigates to the Resting area and rests for 1 minute. When the resting time is over, the robot samples from the entrance probability again to determine if it should continue to rest or to re-enter the tunnel to dig (**Figure 2**).

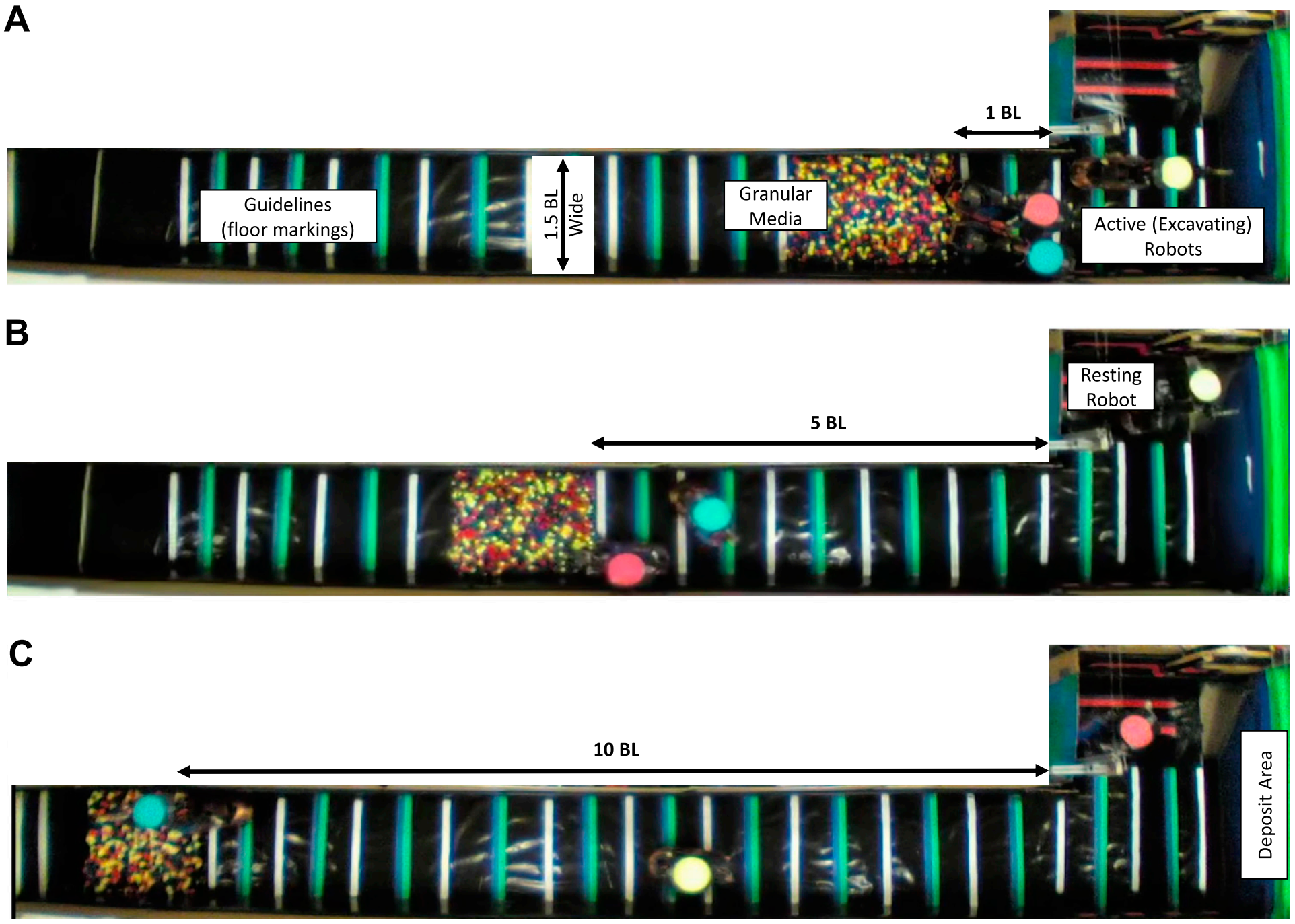

**FIGURE 4 |** Three snapshots of the top view of multi-robot excavation experiments. **(A)** Pellets at one body-length (1BL) of the tunnel which is the initial position of the pellets at the beginning of each trial, **(B)** pellets at five body-length (5BL) of the tunnel after robots have excavated 1200 g of pellets, **(C)** pellets at ten body-length (10BL) of the tunnel after robots have excavated 2700 g of pellets.

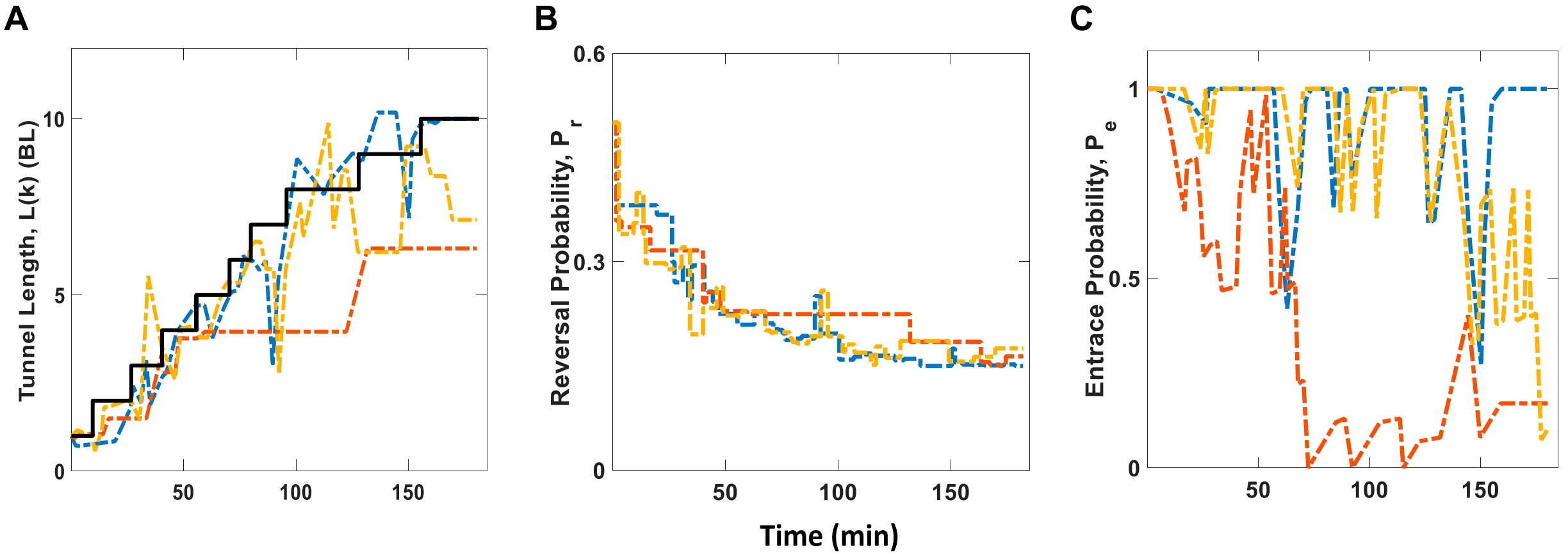

**FIGURE 5 |** Estimated tunnel length, $L(k)$, Reversal probability, $P_r$, and Entrance probability, $P_e$ over time for individual robots based on Adaptive rule. Dashed colored lines represent robot A (red), robot B (blue), and robot C (yellow). Solid black line represents actual tunnel length measurement **(A)** Plot shows how individual robots track the actual growing tunnel, based on dead-reckoning method and exponential moving average formula described in **Eq. 3. (B)** Reversal probability values for individual robots as a function of time. These curves closely match the curve described in **Figure 3B (C)** Plot of the entrance probability values over time. This shows how robots initially started off with equal digging "desires" and gradually become unequal "desires" as the tunnel length grows.

For the Reversal protocol, a fixed reversal probability of 0.8 was used for all the robots in all trials. Prior multi-robot experiments demonstrated that such a high reversal probability regulated congestion better than a lower value. For the Adaptive protocol, however, it is desired that the reversal probability drops rapidly for tunnels less than 5BL and saturates quickly for tunnels greater than 5BL. The value of the $\alpha$ parameter–which controls the maximum and minimum values of the reversal probability for the robot experiment–is set to 0.6 which is within the range of values suggested by the parameter sweep plot of **Figure 3A**.

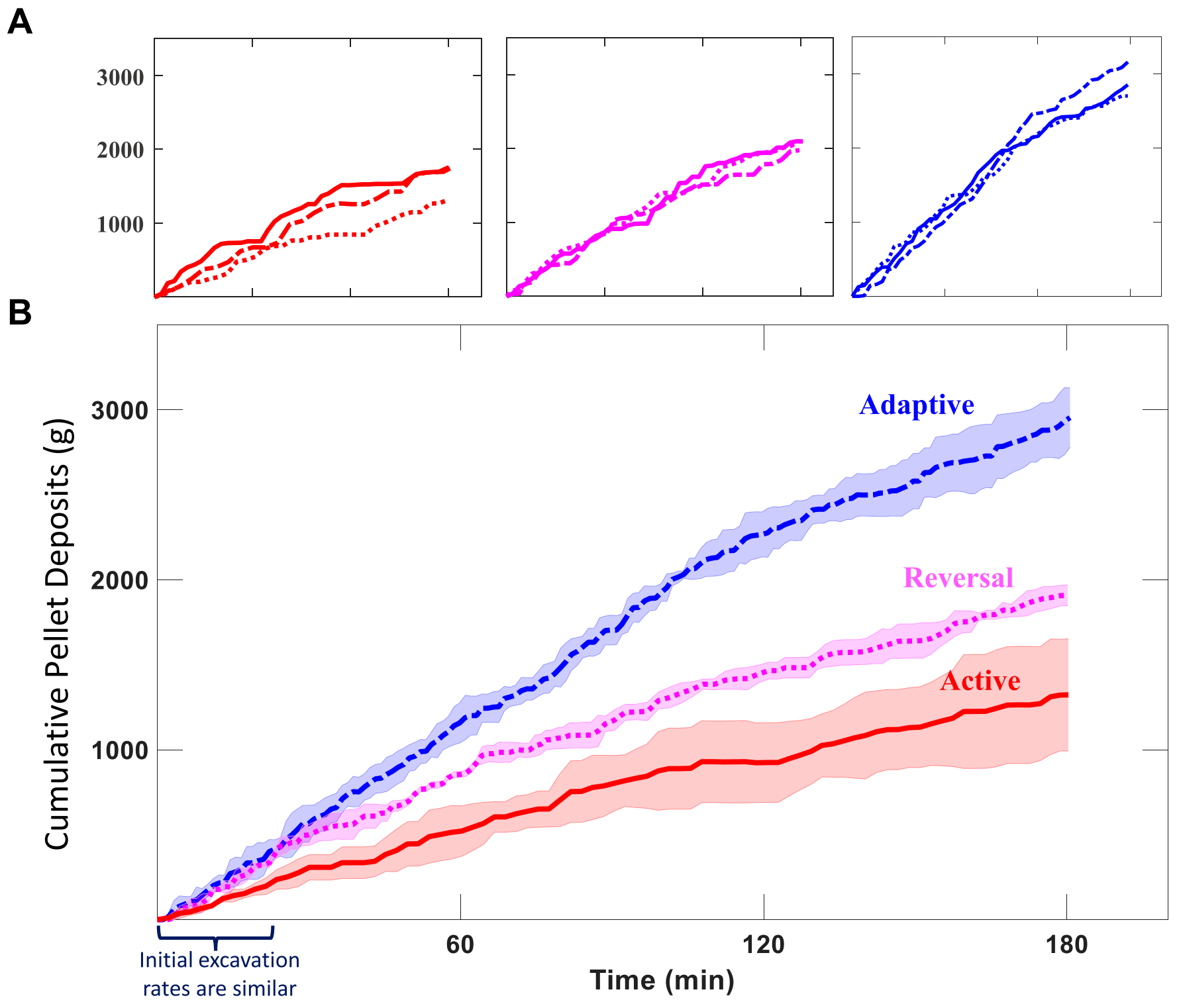

**FIGURE 6 |** Excavated Pellets vs. Time for three different protocols. Active protocol: all robots have an entrance probability of 1.0 but a reversal probability of 0; Reversal protocol: all robots have an entrance probability of 1.0 and reversal probability of 0.8; Adaptive rule: entrance probability is a function of inverse square of tunnel length, while reversal probability is a function of inverse square-root of tunnel length. **(A)** Individual trial comparison of excavation experiments **(B)** Mean excavated pellets, shaded areas correspond to standard deviation from three experiments.

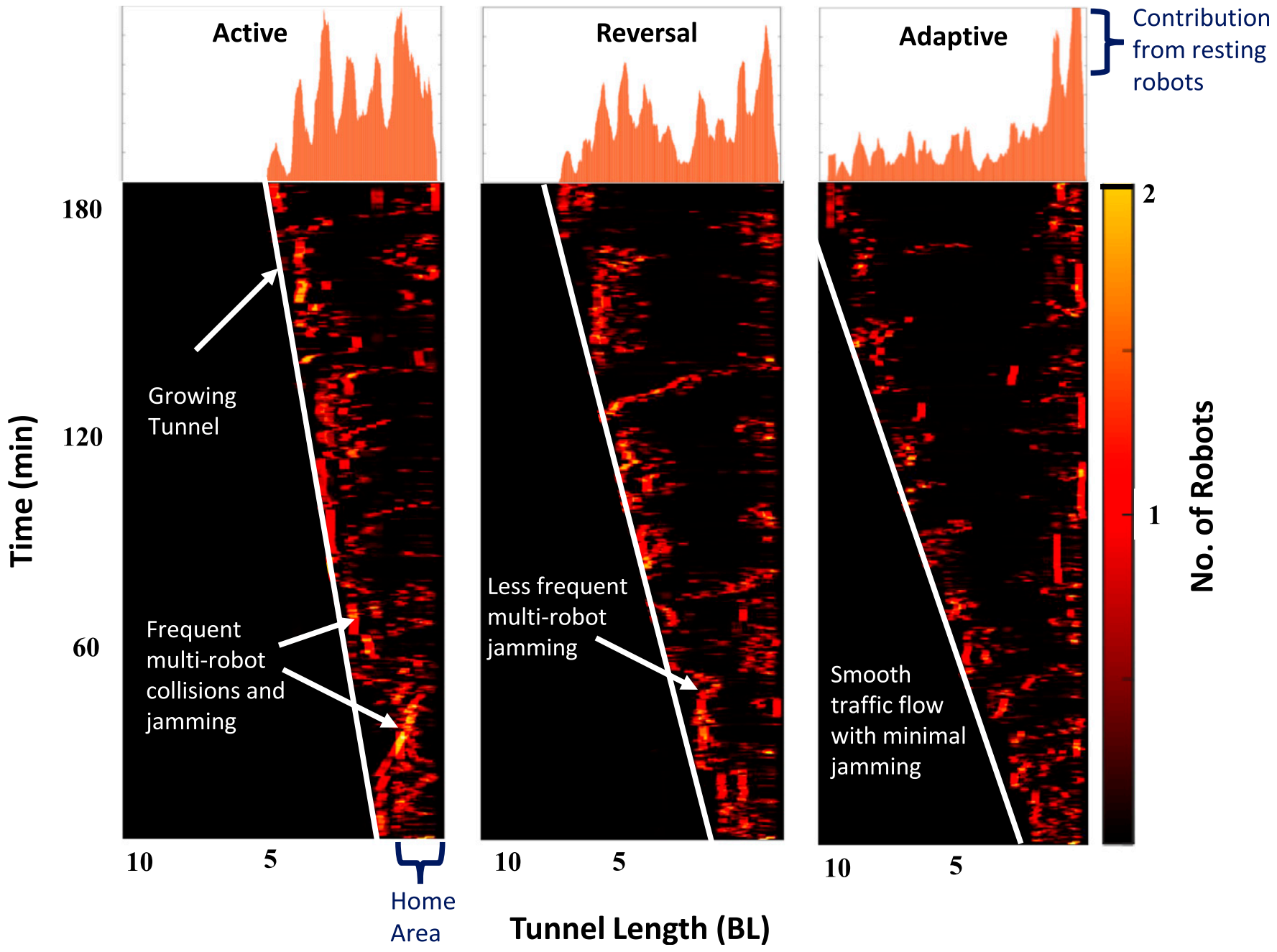

**FIGURE 7 |** Experimental space-time overlap heat maps of robot positions along the tunnel (x-axis) measured in body-lengths (BL). Y-axis is the time duration of the experiment in minutes. White straight lines show how fast the tunnel grows which depends on the running protocol in the robots. Robots start from the Home area (right side) and transit to the Digging area (left side) continuously while excavating the pellets.

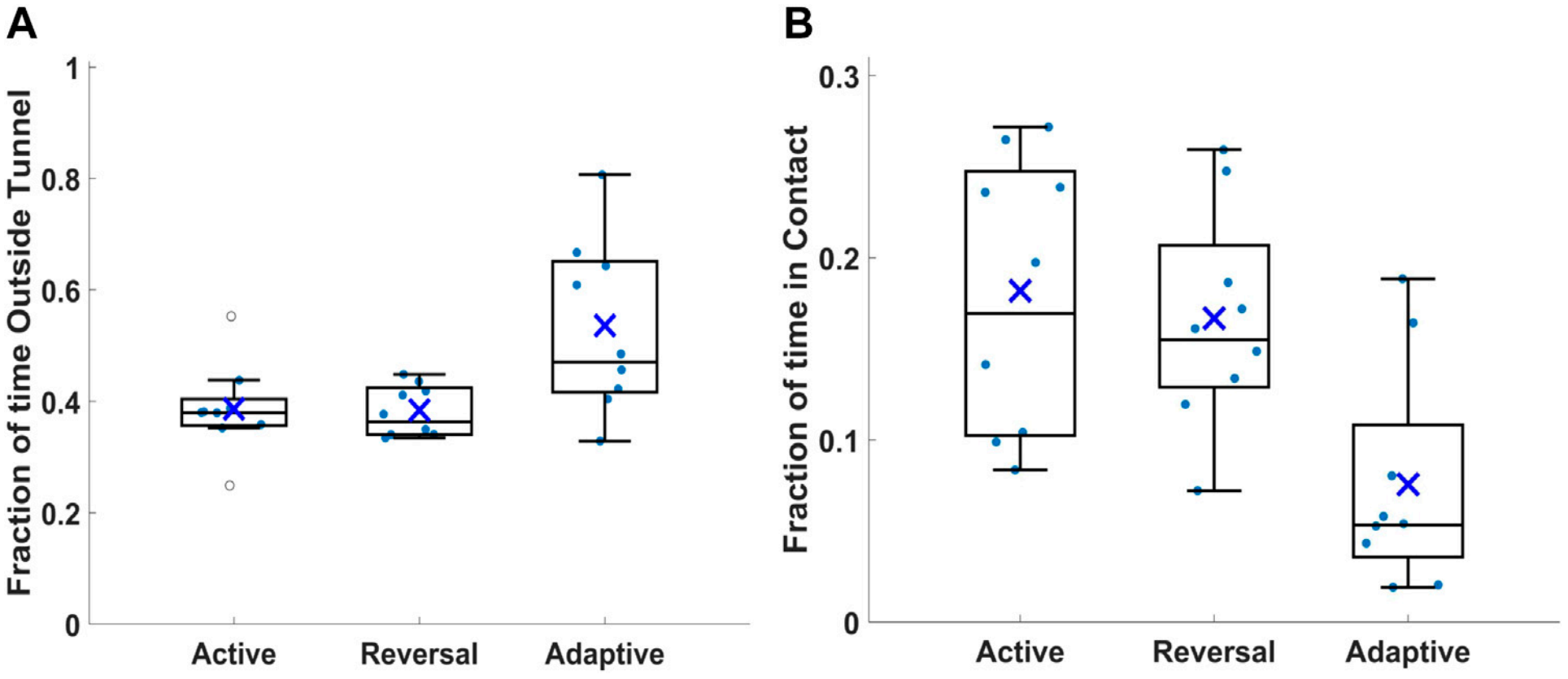

**FIGURE 8 |** Time spent in various states by individual robots, for the three protocols: Active, Reversal and Adaptive. **(A)** Time spent at the resting and deposition areas **(B)** Time spent in contact with other robots while in the tunnel. Boxes span 25th to 75th percentile, whiskers are between the minimal and maximal values, inner line is the median, × for means, and circles (solid and empty) represent actual data points.

# 4 RESULTS

In our robot collectives, we implemented the three different protocols and ran three trials for each protocol. Each trial was conducted for 3 h; this duration is set by the capacity on the power-pack of the robots and ensures sensors and actuators are running effectively. The results of these trials are summarized in **Figures 6–9**.

**Figure 5** shows the estimated tunnel length, $L(k)$, reversal probabilities, $P_r$, and entrance probabilities, $P_e$, for individual robots based on Algorithm 1. **Figure 5A** is the plot of $L(k)$ vs. time which the robots use to estimate the change in tunnel length, ΔL, according to **Eqs 3**, **4**. Since new estimates of tunnel length, $L(k)'$, are derived from the x-component of an individual robot's odometry, it is important to note that wheel slippage can occur when a robot is simultaneously turning and undergoing a collision. This will likely introduce noise in the estimate of the tunnel length, as shown by the fluctuations in the plots of **Figure 5A**. However, the moving average formula of **Eq. 3** will ensure its effect is minimized. In addition, the propagation of noisy measurements is minimized by having the robots reset their odometry measurements at the beginning of each trip, i.e. just before a robot re-enters the tunnel. **Figure 5B** shows that the reversal probability tracks the desired power law expression of **Eq. 1** and **Figure 3B**. **Figure 5C** illustrates how the $P_e$ or "digging desires" of each robot changes from equal to unequal as a function of time, or equivalently, change in tunnel length.

**Figure 6** shows a comparison of the cumulative amount of pellets deposited for the three protocols. **Figure 6A** illustrates that the individual trials with the Adaptive rule yield the highest number of pellet deposits for all trials. The graph shows that all protocols produce similar excavation rates at the initial stages of the experiment before they start to diverge as the tunnel length increases. This confirms that the all protocols and trials started with the same initial conditions, except for the reversal probability values in the case of Adaptive and Reversal protocol.

**Figure 7** shows space-time plots of the robot trajectories for one of the three trials. The presence of robots in the tunnel is tracked from video captured by a camera positioned above the tunnel. At each time point, the presence of robots is summed over the width of the tunnel and is represented by a single row in the diagram. The adaptive rule produces the fastest tunnel growth, and the map includes some stationary blocks near the Home area which corresponds to resting robots.

**Figure 8** explores the portion of time spent by the individual robots either outside of the tunnel (**Figure 8A**)—in the resting or deposition areas–or in contact with each other, while in the tunnel (**Figure 8B**), for the three protocols. The times are quantified based on the robots horizontal position as tracked in the recorded experiments. In **Figure 8B**, robots are considered in contact with others when the horizontal coordinates are less than a body-length apart. We observe that the average time spent outside of the tunnel is roughly the same for the Active and Reversal protocols, at about 40%, but increases a bit for the Adaptive rule (**Figure 8A**). Notably, the Adaptive rule generates a wider variance, indicating some of the robots spend significantly less time in the tunnel than others. Looking at the time spent in contact (**Figure 8B**), we see a narrower distribution for the Reversal than the Active protocol, demonstrating that the Reversal protocol regulates contact time in most cases. More importantly, there is a clear reduction in the average portion of time spent in contact, using the Adaptive rule, from more than 15% using the other protocols, to about 5%.

**Figure 9** compares inequality in workload distribution for the Reversal and Adaptive rules, quantified using Lorenz curves. A Lorenz curve presents the cumulative fraction of work done by a cumulative fraction of the population. This curve is convex by definition and an equal workload distribution appears as a straight line between (0,0) and (1,1) A divergence from this straight line indicates unequal workload distribution, where, for example, half of the population is doing less than half of the work. This measure of divergence is usually quantified by the Gini coefficient, $G$, defined as the ratio of the area between the

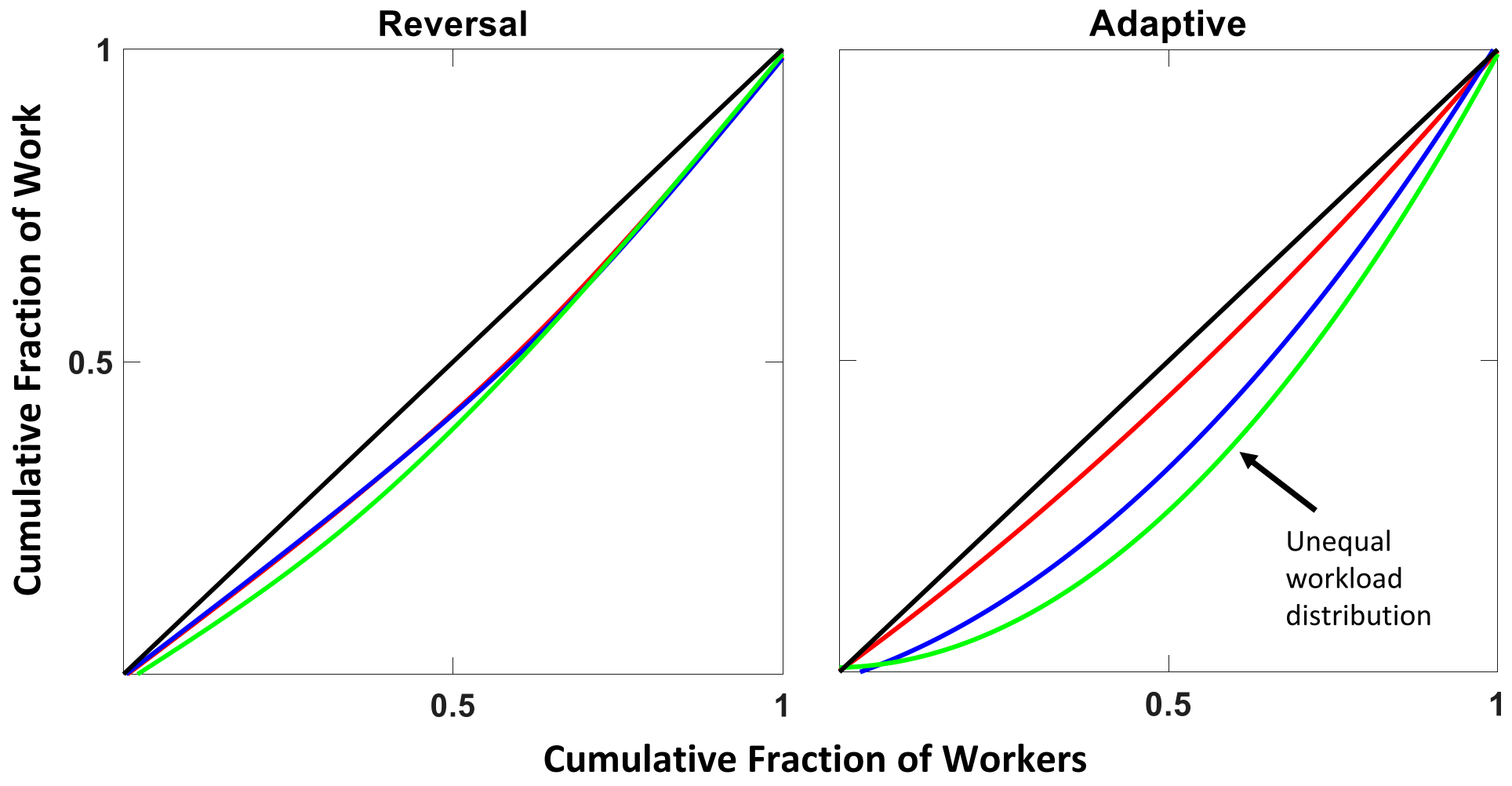

**FIGURE 9 |** Lorenz curves of robot workload distribution for 1-h "epochs" of 3-h experiment. The curves represent: 1 h duration (red), 2 h duration (blue), and 3 h duration (green). Black solid line represents the line of equality. Cumulative fraction of work is the number of trips the robots took into the tunnel in an attempt to excavate pellets. In Reversal protocol, all robots are mostly inside the tunnel, either going in to excavate pellets or going out to dump or "give-up" and re-enter the tunnel. Adaptive rule on the other hand leads some robots to rest and become less active over time.

Lorenz curve and the line of equality [33]. The curves in **Figure 9** show that the Reversal protocol produces an equal workload distribution with a Gini coefficient of approximately ~0.06. The Adaptive protocol on the other hand produces a strategy that leads to equal workload distribution at short tunnels and unequal workload distribution at long tunnels with a Gini coefficient of ~0.3. This strategy produces the most effective excavation rate in all the experiments. This is smaller than the inequality, in terms of a Gini coefficient, previously reported for ants, of about 0.6 [12], which is due to the larger number of individuals involved in the ant study. Since a tunnel essentially imposes a limit on the number of robots or ants that move through it concurrently without clogging, a larger number of individuals requires a higher degree of inequality to avoid clogging.

## 5 DISCUSSION

Our results demonstrate that an adaptive strategy, inspired by observations on ant behavior, leads to significant improvements in performance of excavation through a narrow tunnel, by a group of robots.

It was noticed previously that ants are sometimes willing to reverse or "give up", when faced with oncoming traffic [12]. When studied systematically, it was suggested that there is an optimal probabilistic rate for these reversals, which reduces multi-body collisions and jamming events [12]. Indeed, when we implemented probabilistic reversals upon collisions in the robots, we saw improved performance compared to an insistent, non-reversing ("active") behavior (**Figure 6**). Furthermore, our Cellular Automata simulations of the robots suggest that the optimal reversal rate decreases with increasing tunnel length (**Figure 3**), which results from an increase in time wasted working without achieving pellet excavation. When a jam occurs far into the tunnel, a low reversal probability tells the robot not to give up quickly but rather try to resolve congestion locally.

Despite an improvement in performance (**Figure 6**), the willingness to "give up" and reverse did not significantly reduce contacts between the robots in the system when compared with the "active" protocol (**Figure 8**). Ants display another salient collective feature–an unequal workload distribution–which has been demonstrated to improve performance of collective digging in simulation, when compared to an equal workload distribution [12]. We hypothesized that a reinforcement rule employed by the individual robots, governing entrance probabilities $P_e$ (**Figure 2**), could spontaneously result in an unequal workload distribution.

We implemented a reinforcement rule that increases (decreases) the probability to attempt digging with every successful (unsuccessful) digging trip. This reinforcement rule indeed results in the spontaneous formation of unequal participation in digging (**Figure 9**). Our preliminary experiments showed this unequal workload results in reduced performance for short tunnels and increased performance for long tunnels. Taken together with the trend we observed in simulations for optimal reversal rates (**Figure 3**), we decided to implement adaptive rules employed by individual robots, according to the tunnel length, estimated by distance travelled. As a result, an unequal workload distribution emerged that allows them to avoid costly contacts (**Figure 8**) and collectively perform better in an excavation task (**Figure 6**), using noisy estimates (**Figure 5**).

### 5.1 Relations to Social Insect Collective Dynamics, Active Matter Physics and Swarm Robotics

Given that our work touches on aspects of biological collective behavior, active matter physics and swarm robotics, we briefly discuss our work in context of these well developed disciplines.

In terms of relation to biology, several studies have revealed that social insects (e.g. ants) modify their individual behaviors in response to specific stimuli experienced in the environment [34–38]. This tendency of individuals to make decisions based on their experience or observation is necessary for organisms' survival and reproduction, and it is termed adaptation or learning [39, 40]. In ants removal of highly active ants from the group results in increased activity by the others [41], suggesting they too use some adaptive strategy. Other studies have also observed adaptation in ant collectives. For example, Buhl et al [42] suggested a feedback model that explains the excavation behavior of ants in a laboratory setting, and Bruce et al [27] suggested that ants use collision information to maintain a desired proximity to others. Thus, we find it reasonable that ants adapt their behaviors based on excavation success which could have a strong relation to collisions (traffic jams) and tunnel length.

Biological systems are known to possess compliant and flexible capabilities which enable them to perform sophisticated maneuvers that are otherwise difficult for their robotic counterparts [43]. For example, Gravish et al [44] studied how antenna deformations provide mechanical support to slipping ants when climbing in confined spaces. Such morphological adaptation of ants makes them excellent excavators in their natural environments [45]. Ants typically generate tunnels that fit about two ant widths and can easily pass each other within them. However, an encounter with three or more ants will take longer to resolve. Our robots also take much longer to resolve 3-robot traffic jams than they do 2-robot collisions, and these become the dominant time cost to be avoided. Thus, strategies for congestion modulation in biological systems may prove applicable in multi-robot real-world scenarios.

In terms of relation of our work to active matter physics, most studies of active matter assume particles remain in a given state (e.g., constant speed movement) and study the global dynamics emerging from such rules. There is typically no "goal" for the global dynamics in such studies. In contrast our system is explicitly "task oriented": from a broad perspective, the system has a mixture of particles with different behaviors that occasionally transition between the different populations (control states) in pursuance of a goal. Thus from a physics perspective it is interesting to ask how desired macroscopic outcomes (e.g., flow of material) must be coupled to microscopic rules which can change in response to macroscopic state (e.g., particles "give up" which detect a slowing of flow). This is particularly interesting in the collisional and dense regimes in part because of challenges for any one agent to know the state of others and active systems must deal with the propensity of such systems to cluster [9, 46, 47], clog and form glassy states [11].

Finally, from a swarm robotics/engineering point of view, while tasks are a critical aspect of making swarms task capable, most work has been conducted in either low density regimes and/or focused on various techniques for collision avoidance [48–52]. This is to ensure safe operation of the robots and to prevent possible catastrophes that may occur in cases of collisions. However, attempting to avoid collisions in crowded and confined conditions could be impractical. This is due to the physical constraint of the environment and/or the uncertainties present in sensor measurements which make collisions inevitable. Even in some cases where accurate measurements are available, the challenges in such environments make robot to be overtly cautious. This conservative behavior would make the robots spend most of their time avoiding collisions rather than advancing the mission of the group.

Recently, researchers have studied scenarios involving small mass and low velocity robots where mild collisions and contacts can be tolerated. In this case, collision can be used as a sensing modality to estimate the state of the environment. For example, [53, 54] developed a probabilistic filtering technique based on inter-robot contacts to localize a team of robots in particular environments. The robots were equipped with binary tactile collision sensors, which provide information for computing the likelihood of a robot to experience collisions in different sections of the environment. In contrast to this, our work uses collisions as a source of reinforcement rather than to estimate the state of the environment. We assume the environment is unknown and dynamic, so our approach can generalize to various scenarios where little domain knowledge is available. This requires our robots to learn to cooperate and adjust to the changes in the environment, including the behavior of other robots, to accomplish their tasks collectively and effectively.

# 6 CONCLUSION

Active particles performing persistent motion often develop structures consisting of aggregated formations [3–5], especially in confined environments [10–12]. Active matter studies typically involve particles lacking sensing and control that change direction only stochastically. Our robotic system presents a kind of task-oriented active matter, where control is injected to minimize otherwise unavoidable aggregated states and improve collective performance. In this work, collective performance is measured as the rate of pellet collection from the tip of an ever-extending tunnel. This imposed task compels our robots to traverse the entire length of the system (tunnel), to and fro, while encountering any other robot already in it.

Unlike uncontrolled particles, the robots studied here do not stochastically change direction. In fact, when impeded by another robot, they first attempted to maneuver around the robot to continue on their way. This behavior facilitates a "greedy" attempt to maximize individual performance, at the expense of the collective one, and exacerbates formation of aggregates (**Figure 7** and **Figure 8**). On the other hand, we demonstrated that robots can limit time wasted on this persistence through probabilistic reversal, which generates some collective performance gains (**Figure 6**). However, aggregates still form and are expected to increase in frequency with collectives larger than those studied here.

Even in our relatively small groups of robots, to achieve high collective performance that is robust to a changing environment and possibly to group size, we developed a learning control scheme that uses collisions as an information source, which are a noisy proxy for number density in the system. In robotics, collisions are often viewed as problematic occurrences to be avoided, whereas, they could be an important aspect in the lives of social insects, given the constraints and challenges of their environments. Collisions could serve as fundamentally important information sources that can be harnessed to coordinate the activities of individuals so as to achieve the common goal of the group.

In our multi-robot scenario, we achieved coordination for effective excavation performance by modifying individual robot's response to collision and task desires based on an independent estimate of the tunnel length. We discovered that "giving up", while sacrificing the individual performance, often contributes to the collective performance. We demonstrated that a learning rule that modulates both "giving up" rate and "individual desires" gives a significantly higher group performance than with maladaptive behaviors (**Figure 6**). This technique could be applied to real world scenarios where collisions or physical interactions are unavoidable, or in decentralized task-oriented physical systems where individuals in a group must interact via contacts. We suspect social insects also make good use of collisions to modulate their decision making, in service of a collective goal.

## DATA AVAILABILITY STATEMENT

The raw data supporting the conclusions of this article will be made available by the authors, without undue reservation.

## AUTHOR CONTRIBUTIONS

KA, H-SK, MB, and DG devised the project, algorithm and experiments. KA developed the robots, algorithms and performed the experiments. KA and RA performed data analysis. KA, RA, MG, and DG contributed to the writing of the manuscript. H-SK, MB, MG, and DG provided insightful discussions and support.

## FUNDING

Funding for this research provided by ARO MURI award W911NF-19-1-023, NSF Grant PHY-1205878, NSF Grant IOS-2019799 to MG and DG, and NSF Grant DMR-1725065 to MB.

## ACKNOWLEDGMENTS

We are grateful to Andrea W. Richa and Joshua Daymude of Arizona State University for their helpful discussions.

## SUPPLEMENTARY MATERIAL

The Supplementary Material for this article can be found online at: https://www.frontiersin.org/articles/10.3389/fphy.2022.735667/full#supplementary-material

**Conflict of Interest:** The authors declare that the research was conducted in the absence of any commercial or financial relationships that could be construed as a potential conflict of interest.

**Publisher's Note:** All claims expressed in this article are solely those of the authors and do not necessarily represent those of their affiliated organizations, or those of the publisher, the editors and the reviewers. Any product that may be evaluated in this article, or claim that may be made by its manufacturer, is not guaranteed or endorsed by the publisher.